\documentclass[letterpaper]{article} 
\usepackage{aaai2027}  
\usepackage[hyphens]{url}  
\usepackage{graphicx} 
\usepackage{natbib}  
\usepackage{caption} 
\usepackage{algorithm}
\usepackage{algorithmic}

\usepackage{newfloat}
\usepackage{multirow}
\usepackage{amsmath}
\usepackage{amsfonts}
\usepackage{listings}
\DeclareCaptionStyle{ruled}{labelfont=normalfont,labelsep=colon,strut=off} 
\floatstyle{ruled}
\newfloat{listing}{tb}{lst}{}
\floatname{listing}{Listing}

\usepackage{booktabs}

\title{CoRe-UIE: Rethinking Coexisting and Region-wise Degradation for Underwater Image Enhancement}
\author{
Weifeng Kong\textsuperscript{\rm 1},
Chenghao Xu\textsuperscript{\rm 1},
Lin Chen\textsuperscript{\rm 1},
Ziheng Cao\textsuperscript{\rm 1}
Guanying Huo \textsuperscript{\rm 1\corresponding},
}

\affiliations{
\textsuperscript{\rm 1}Hohai University\\
Jiangsu, China\\
}

\begin{document}

\maketitle

\begin{abstract}
Underwater images often suffer from diverse and coexisting degradations, including color distortion, scattering haze, texture attenuation, and uneven illumination. These degradations vary across regions and may coexist locally, making conventional uniform restoration difficult to adapt to different degradation patterns. To address this problem, we propose Coexisting and Region-wise Degradation for Underwater Image Enhancement (\textbf{CoRe-UIE}), a degradation-oriented expert collaboration framework. CoRe-UIE combines a content-preserving shared expert with four shared-backbone routed experts for color correction, scattering suppression, texture recovery, and illumination protection. The routed experts share the same architecture but have independent parameters, and are assigned to different regions through input-derived degradation cues and region-adaptive Top-\(k\) routing. We further introduce a Hilbert--Schmidt Independence Criterion (HSIC)-based representation constraint to reduce statistical dependence among expert features and alleviate redundant expert responses. Experiments on UIEB, LSUI, and U45 demonstrate that CoRe-UIE achieves competitive quantitative performance and visually balanced enhancement under diverse underwater degradation conditions.
\end{abstract}

\section{Introduction}

\begin{figure}
    \centering
    \includegraphics[width=1\linewidth]{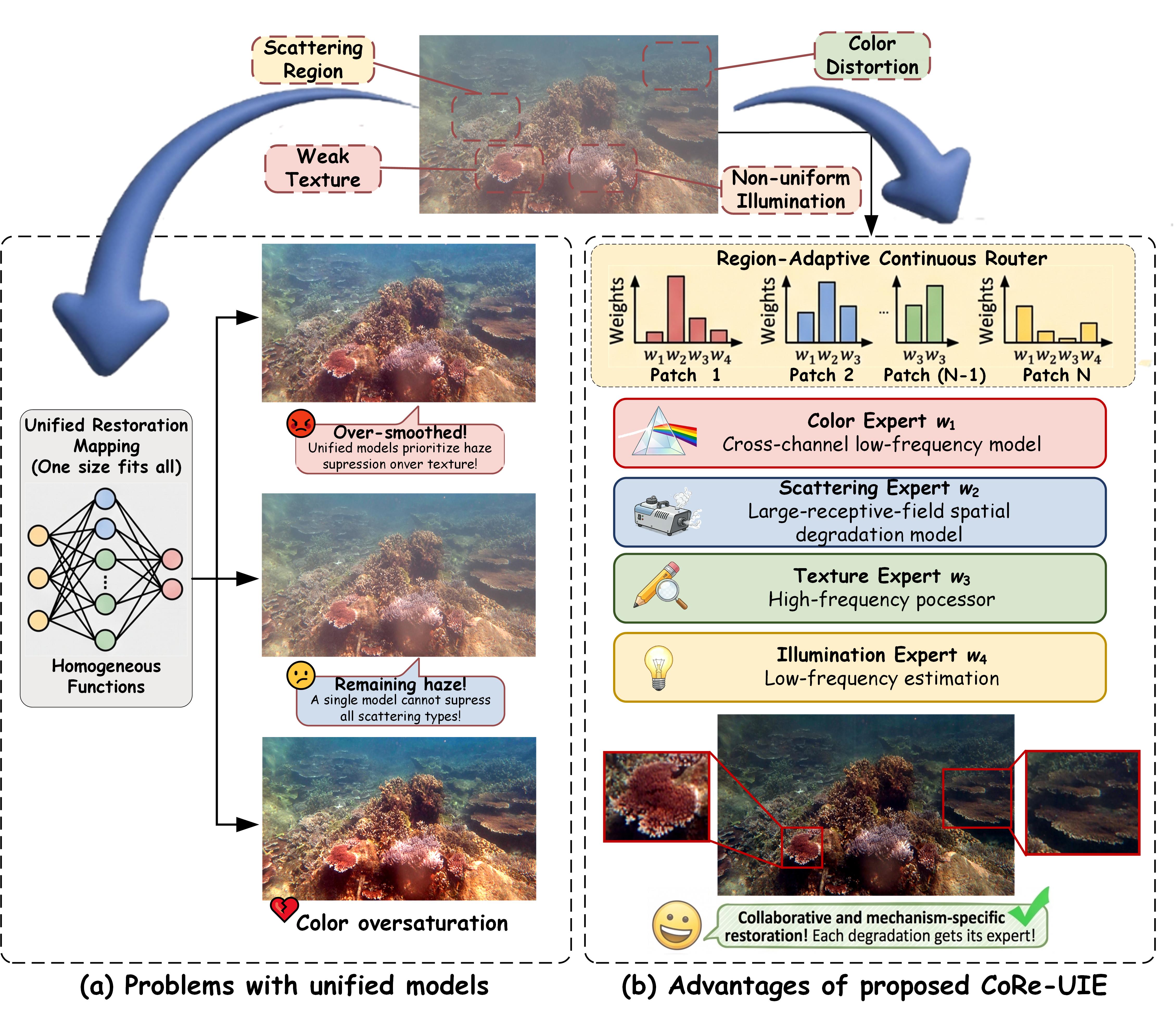}
    \caption{Motivation of CoRe-UIE.}
    \label{fig:motivation}
\end{figure}

Underwater imaging plays an important role in marine exploration, ecological monitoring, and underwater robotic perception, where image enhancement and restoration are often required to support reliable visual analysis under degraded underwater conditions~\cite{zhu2026efficient,yu2026task,zhu2026ueaod,wen2026joint}. However, underwater degradation is spatially heterogeneous and mechanism-compositional: color distortion, scattering haze, texture degradation, and uneven illumination may appear with different strengths across regions and even coexist within the same local area. Existing unified restoration networks usually process all regions through a shared mapping, which may lead to residual color casts, incomplete haze removal, over-smoothed textures, or illumination artifacts. As shown in Fig.~\ref{fig:motivation}, underwater enhancement therefore requires region-adaptive collaboration among different restoration mechanisms rather than a one-size-fits-all mapping.

Previous underwater enhancement methods have explored physical imaging models, handcrafted priors, and multi-scale fusion to compensate for wavelength-dependent attenuation and scattering~\cite{akkaynak2019seathru,li2019uieb}. However, their performance is often sensitive to assumptions about water quality, ambient illumination, and transmission estimation. Recent learning-based methods further improve restoration quality with more expressive network designs~\cite{Cong2026review, ANWAR2020115978}. For example, Ucolor exploits multi-color-space representations and transmission guidance~\cite{li2021ucolor}, PUIE-Net considers the uncertainty of imperfect reference images~\cite{fu2022puie}, Spectroformer enhances spectral and spatial modeling with Transformer architectures~\cite{khan2024spectroformer}, WF-Diff introduces frequency-domain diffusion for underwater restoration~\cite{zhao2024wfdiff}, and WaterMamba and O-Mamba employ state-space models for efficient long-range dependency modeling~\cite{guan2024watermamba,dong2024omamba}. Although these methods have significantly advanced underwater image enhancement, they mainly improve feature representation, degradation modeling, or restoration capacity within a general enhancement framework. The mechanism-compositional nature of underwater degradation, where color distortion, scattering haze, texture degradation, and illumination variation may appear with different strengths across regions, is still not explicitly modeled. Consequently, existing methods provide limited flexibility for region-dependent and mechanism-specific correction.

Mixture-of-experts (MoE) architectures enhance model capacity by routing features to different expert subnetworks~\cite{shazeer2017moe,lepikhin2020gshard,fedus2021switch}. Visual MoE further extends this idea to image tokens through sparse expert routing~\cite{riquelme2021vmoe}, while Soft MoE improves optimization stability with differentiable soft dispatching and aggregation~\cite{puigcerver2024softmoe}. However, conventional Top-\(k\) MoE methods usually employ homogeneous experts whose specialization is learned implicitly from data. This design is insufficient for underwater restoration, where color distortion, scattering, texture degradation, and illumination variation vary across regions and may coexist locally. Without mechanism-aware expert design, different experts may learn redundant corrections and produce routing decisions with limited restoration semantics. Therefore, an effective MoE-based underwater enhancement model should combine adaptive routing with mechanism-aware expert regularization, so that experts with comparable structures can still learn interpretable and complementary restoration behaviors.

To address these limitations, we propose \textbf{CoRe-UIE}, an expert collaboration framework for underwater image enhancement. As shown in Fig.~\ref{fig:motivation}(b), CoRe-UIE consists of a content-preserving shared expert, a set of shared-backbone routed experts, and a region-adaptive router. The shared expert captures common restoration knowledge that should be preserved across different underwater regions, while the routed experts adopt the same backbone architecture with independent parameters. Instead of manually designing different expert structures for different degradation types, CoRe-UIE encourages expert specialization through degradation-related cues and routing constraints. Specifically, the router integrates learnable degradation representations with input-derived physical cues, such as color imbalance, scattering-related contrast degradation, texture structure, and illumination risk, and assigns spatially varying weights to the most relevant experts. In this way, different regions can be restored by different expert combinations according to their local degradation characteristics. We further introduce mechanism-guided response alignment and HSIC-based representation disentanglement to encourage degradation-consistent expert responses and reduce statistical dependence among expert representations~\cite{gretton2005hsic}.

Our main contributions are summarized as follows:
\begin{itemize}
    \item We propose \textbf{CoRe-UIE}, a mechanism-aware adaptive restoration framework for underwater image enhancement, which addresses spatially diverse and locally coexisting underwater degradations through shared-backbone expert collaboration.

    \item We design a region-adaptive Top-\(k\) routing strategy that selects different expert combinations for different image regions according to local degradation characteristics, enabling adaptive restoration under mixed underwater degradation patterns.

    \item We introduce a mechanism-cue-guided expert specialization objective that combines response alignment and HSIC-based representation disentanglement, encouraging experts with the same architecture to learn less redundant and more complementary restoration representations.
\end{itemize}

\section{Related Work}

\subsection{Underwater Image Enhancement}

Early underwater image enhancement methods mainly relied on physical imaging models and handcrafted priors. Sea-Thru~\cite{akkaynak2019seathru} explicitly models wavelength-dependent attenuation and backscatter, while UIEB~\cite{li2019uieb} provides a widely used benchmark for evaluating enhancement methods. However, physics-based approaches often depend on depth information, scene geometry, or assumptions about water properties, which limits their robustness in diverse real-world environments. Recent learning-based methods have shifted underwater enhancement toward data-driven restoration: U-shape Transformer~\cite{peng2023ushape} combines multi-scale fusion with global dependency modeling, Semi-UIR~\cite{huang2023semiuir} reduces paired-data dependence through semi-supervised contrastive learning, and Five A$^{+}$ Network~\cite{jiang2023fiveaplus} explores compact and efficient enhancement. Although these methods improve restoration quality, supervision flexibility, or efficiency, they still mainly operate within a general enhancement framework. The region-dependent and mechanism-compositional nature of underwater degradation remains less explicitly explored, where color distortion, scattering, texture loss, and illumination variation may appear with different strengths across regions and coexist locally. In contrast, CoRe-UIE introduces a content-preserving shared expert and four structurally differentiated routed experts for color correction, scattering suppression, texture recovery, and illumination protection.

\subsection{Mixture-of-Experts and Mechanism-Aware Restoration}

MoE architectures increase model capacity through conditional computation. Sparsely-gated MoE~\cite{shazeer2017moe} introduced trainable gating to activate a small subset of experts, while GShard~\cite{lepikhin2020gshard} and Switch Transformer~\cite{fedus2021switch} improved the scalability and efficiency of sparse expert models. In vision, V-MoE~\cite{riquelme2021vmoe} routed image tokens to expert networks, and Soft MoE~\cite{puigcerver2024softmoe} further improved optimization stability with differentiable soft dispatching.

MoE has also been explored for low-level vision and image restoration, where experts are organized according to task type, degradation category, image content, or restoration difficulty. For example, MoCE-IR~\cite{zamfir2025moceir} assigns experts according to image complexity for adaptive restoration. Although these methods demonstrate the flexibility of conditional expert collaboration, expert specialization is often learned implicitly from data, and its relationship with degradation mechanisms is not always clear. This issue is more challenging in underwater enhancement, where color distortion, scattering haze, texture degradation, and illumination variation are spatially heterogeneous and may coexist locally. Directly applying conventional MoE may lead to redundant expert responses or ambiguous routing semantics. In contrast, CoRe-UIE adopts shared-backbone routed experts and encourages their functional differentiation through input-derived mechanism cues, region-adaptive routing, response alignment, and HSIC-based representation disentanglement, enabling mechanism-aware collaboration without manually crafted heterogeneous expert structures.

\section{Methodology}

\subsection{Overview}

\begin{figure*}[t]
    \centering
    \includegraphics[width=1\linewidth]{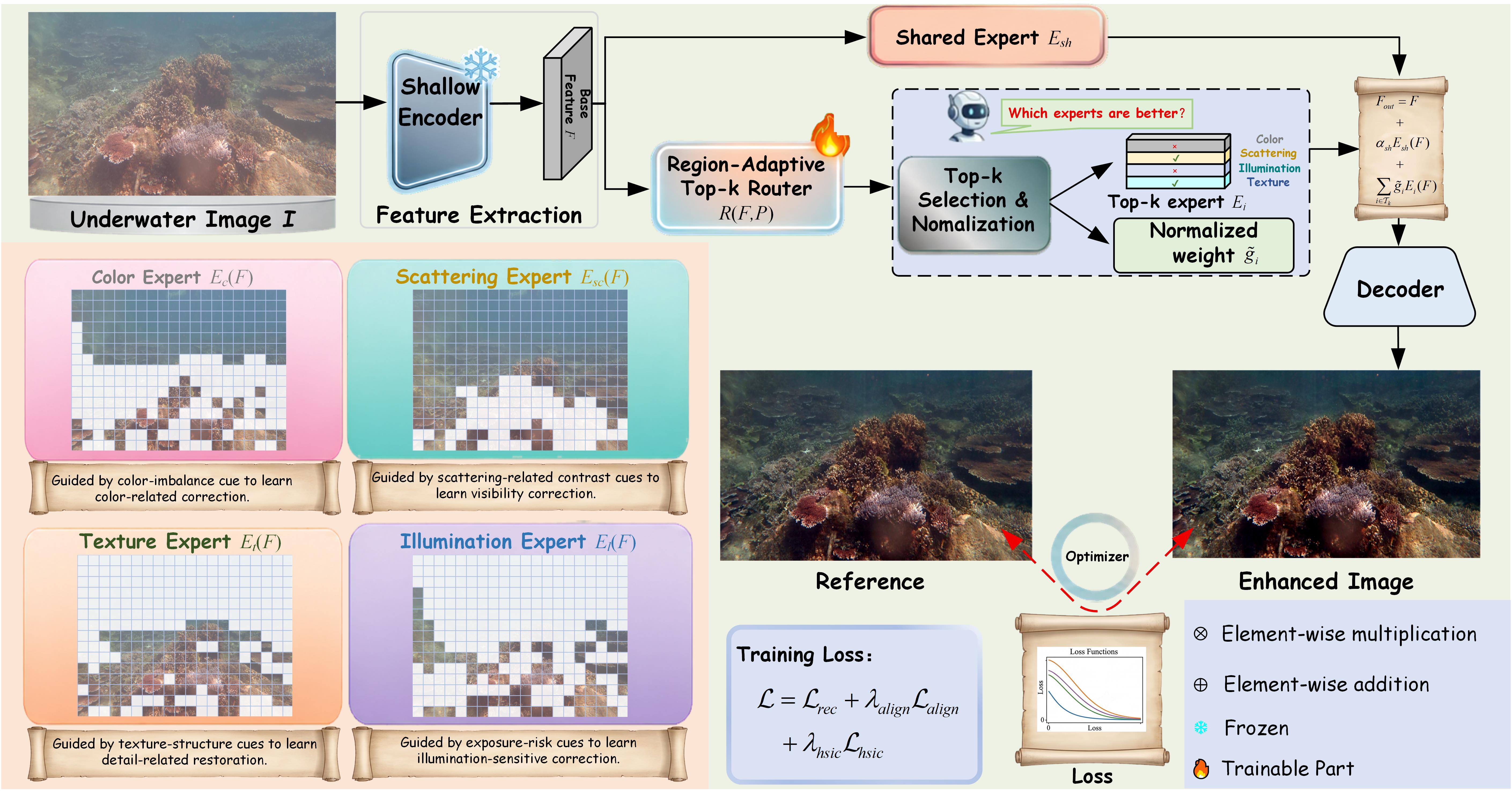}
    \caption{Overall architecture of the proposed CoRe-UIE framework.}
    \label{fig:framework}
\end{figure*}

As shown in Fig.~\ref{fig:framework}, the proposed CoRe-UIE enhances underwater images through a content-preserving shared branch and a region-adaptive expert collaboration branch. Given an underwater image \(I\in \mathbb{R}^{3\times H\times W}\), a shallow encoder first extracts the base feature representation:
\begin{equation}
F=\phi(I),
\label{eq:encoder}
\end{equation}
where \(I\) denotes the input underwater image, \(\phi(\cdot)\) is the shallow feature encoder, and \(F\in\mathbb{R}^{C\times H'\times W'}\) is the extracted base feature. Here, \(C\) denotes the feature dimension, and \(H'\times W'\) denotes the spatial resolution of the encoded feature. The encoder is implemented with a convolutional stem and residual feature extraction blocks, which project the input image into a high-dimensional feature space.

CoRe-UIE contains one content-preserving shared expert and four shared-backbone routed experts. The shared expert captures degradation-invariant content information, while the routed experts have the same architecture but independent parameters. For clarity, we denote the shared expert as \(E_{sh}\), and the four routed experts as \(E_c\), \(E_{sc}\), \(E_t\), and \(E_l\), corresponding to color-related correction, scattering-related suppression, texture-related recovery, and illumination-related protection, respectively.

A region-adaptive Top-\(k\) router predicts spatially varying expert weights from the encoded feature \(F\) and input-derived mechanism cues \(P\), including color imbalance, scattering-related contrast degradation, texture structure, and illumination risk. The selected expert outputs are fused with the shared expert output and decoded into the enhanced image:
\begin{equation}
\hat{Y}=D(F_{out}),
\label{eq:decoder}
\end{equation}
where \(F_{out}\) is the fused feature, \(D(\cdot)\) denotes the reconstruction decoder, and \(\hat{Y}\) is the enhanced image. During inference, only the underwater input image is required.

During training, the enhanced image \(\hat{Y}\) is compared with the reference image \(Y\) to compute the reconstruction objective, while response alignment and HSIC-based representation disentanglement are used to guide expert specialization. During inference, only the underwater input image is required, and the reference image and loss branches are removed.

\subsection{Shared Expert and Shared-Backbone Mechanism Experts}

CoRe-UIE contains one content-preserving shared expert and four routed mechanism experts. The shared expert is introduced to preserve degradation-invariant visual content, such as scene layout, object boundaries, and basic structural information. It is activated for all spatial regions and provides common restoration features that stabilize the final reconstruction.

For region-dependent correction, CoRe-UIE employs four routed experts corresponding to color-related correction, scattering-related suppression, texture-related recovery, and illumination-related protection. Instead of manually assigning different network structures to different degradation types, all routed experts adopt the same backbone architecture but have independent parameters. This design keeps all experts with comparable modeling capacity, while their functional differences are induced by input-derived mechanism cues, region-adaptive routing, response alignment, and HSIC-based representation disentanglement.

Given the encoded feature \(F\), the \(i\)-th routed expert is implemented as a residual restoration block:
\begin{equation}
E_i(F)
=
F+
\mathcal{C}^{2}_{i}
\left(
\sigma
\left(
\mathcal{C}^{1}_{i}(F)
\right)
\right),
\label{eq:expert_block}
\end{equation}
where \(E_i(F)\) denotes the output of the \(i\)-th routed expert, \(\mathcal{C}^{1}_{i}\) and \(\mathcal{C}^{2}_{i}\) are learnable convolutional transformations, and \(\sigma(\cdot)\) denotes the nonlinear activation function. Although the routed experts share the same block design, their parameters are independent, allowing each expert to learn a different restoration response.

We denote the four routed experts as \(E_c\), \(E_{sc}\), \(E_t\), and \(E_l\), corresponding to color, scattering, texture, and illumination related restoration roles, respectively. These roles are not imposed by hand-crafted architectural differences. Instead, the router assigns experts to different spatial regions according to degradation cues, while the response alignment loss encourages each expert to respond to regions consistent with its corresponding cue. Meanwhile, HSIC-based representation disentanglement reduces statistical dependence among expert features, encouraging experts with the same architecture to learn less redundant restoration representations. In this way, CoRe-UIE obtains mechanism-aware expert collaboration without relying on manually designed heterogeneous expert structures.

\subsection{Region-Adaptive Top-\(k\) Routing}

Since all routed experts adopt the same backbone structure, the router plays a central role in assigning different restoration functions to different image regions. Fixed expert fusion lacks regional flexibility, while single-expert routing is insufficient for underwater scenes where multiple degradation factors may coexist locally. We therefore use a region-adaptive Top-\(k\) router to select the most relevant expert combinations according to local degradation characteristics.

We extract input-derived mechanism cues \(P=\{P_c,P_{sc},P_t,P_l\}\), corresponding to color imbalance, scattering-related contrast degradation, texture structure, and illumination risk, respectively. Given the encoded feature \(F\) and the mechanism cues \(P\), the router predicts the expert weight distribution:
\begin{equation}
G=\mathrm{Softmax}(R(F,P)),
\label{eq:router}
\end{equation}
where \(R(\cdot)\) is the routing network and \(G=\{g_i\}_{i=1}^{N}\) denotes the predicted expert weights.

For each spatial region, the router selects the Top-\(k\) experts with the largest weights:
\begin{equation}
\mathcal{T}_k=\mathrm{TopK}(G),
\label{eq:topk_set}
\end{equation}
and renormalizes their routing weights as:
\begin{equation}
\tilde{g}_i
=
\frac{g_i}
{\sum_{j\in\mathcal{T}_k}g_j},
\quad i\in\mathcal{T}_k .
\label{eq:topk_norm}
\end{equation}
Here, \(g_i\) is the original routing weight of the \(i\)-th expert, and \(\tilde{g}_i\) is the normalized weight used for feature fusion.

The Top-\(k\) strategy allows multiple experts to collaborate within the same local region. Compared with Top-1 routing, it avoids assigning a region to only one restoration mechanism. Compared with dense fusion, it maintains expert selectivity and reduces redundant expert activation. In our implementation, we set \(k=2\) for all experiments.

\subsection{Loss Functions}

Although the region-adaptive Top-\(k\) router assigns different expert combinations to different regions, routing weights alone cannot guarantee clear expert specialization. Therefore, CoRe-UIE is optimized with a joint objective that combines reconstruction fidelity, response alignment, and expert representation disentanglement:
\begin{equation}
\mathcal{L}
=
\mathcal{L}_{rec}
+
\lambda_{align}\mathcal{L}_{align}
+
\lambda_{hsic}\mathcal{L}_{hsic},
\label{eq:total_loss}
\end{equation}
where \(\mathcal{L}_{rec}\) is the reconstruction loss, \(\mathcal{L}_{align}\) guides expert responses with mechanism cues, and \(\mathcal{L}_{hsic}\) reduces redundancy among expert representations. The coefficients \(\lambda_{align}\) and \(\lambda_{hsic}\) control the relative importance of the two regularization terms.

For reconstruction supervision, we follow the common practice in recent image restoration and underwater enhancement methods~\cite{zamir2022restormer,khan2024spectroformer}. The reconstruction loss \(\mathcal{L}_{rec}\) combines pixel-wise fidelity terms with structural consistency constraints, providing the main supervision for recovering images that are close to the reference image in both intensity and structure.

To encourage each expert to focus on regions consistent with its restoration role, we define the spatial response of the \(i\)-th expert using its output feature \(E_i(F)\) and normalized routing weight \(\tilde{g}_i\):
\begin{equation}
R_i
=
\tilde{g}_i |E_i(F)|,
\label{eq:expert_response}
\end{equation}
where \(F\) is the encoded feature and \(|\cdot|\) denotes element-wise absolute value. Let \(P_i\) denote the input-derived mechanism cue corresponding to the \(i\)-th expert, such as color imbalance, scattering haze, texture structure, or illumination risk. The response alignment loss is formulated as:
\begin{equation}
\mathcal{L}_{align}
=
\sum_i
\left\|
\mathrm{Norm}(R_i)
-
\mathrm{Norm}(P_i)
\right\|_1 ,
\label{eq:align_loss}
\end{equation}
where \(\mathrm{Norm}(\cdot)\) denotes spatial normalization. This term softly aligns expert responses with their corresponding degradation-related regions without requiring pixel-level degradation annotations.

Because the routed experts share the same backbone architecture and are optimized under the same reconstruction objective, their intermediate representations may become correlated. To reduce such redundancy, we introduce an expert representation disentanglement loss based on the Hilbert--Schmidt Independence Criterion (HSIC)~\cite{gretton2005hsic}. Given two random variables \(X\) and \(Y\), HSIC measures the squared Hilbert--Schmidt norm of their cross-covariance operator:
\begin{equation}
\mathrm{HSIC}(X,Y)
=
\left\|
\mathcal{C}_{XY}
\right\|_{\mathrm{HS}}^{2},
\label{eq:hsic_operator}
\end{equation}
where \(\mathcal{C}_{XY}\) denotes the cross-covariance operator and \(\|\cdot\|_{\mathrm{HS}}\) is the Hilbert--Schmidt norm.

Let \(Z_i\) and \(Z_j\) denote intermediate representations from the \(i\)-th and \(j\)-th experts. For a mini-batch with \(m\) samples, the empirical HSIC is computed as:
\begin{equation}
\mathrm{HSIC}(Z_i,Z_j)
=
\frac{1}{(m-1)^2}
\mathrm{tr}
\left(
K_i H K_j H
\right),
\label{eq:hsic_empirical}
\end{equation}
where \(K_i\) and \(K_j\) are kernel matrices computed from \(Z_i\) and \(Z_j\), \(H=I-\frac{1}{m}\mathbf{1}\mathbf{1}^{\top}\) is the centering matrix, and \(\mathrm{tr}(\cdot)\) denotes the matrix trace.

The HSIC loss is defined over all pairs of routed experts:
\begin{equation}
\mathcal{L}_{hsic}
=
\frac{1}{N(N-1)}
\sum_{i=1}^{N}
\sum_{\substack{j=1 \\ j\neq i}}^{N}
\mathrm{HSIC}(Z_i,Z_j),
\label{eq:hsic_loss}
\end{equation}
where \(N\) is the number of routed experts. By reducing statistical dependence among expert representations, HSIC encourages experts with the same architecture to learn less redundant restoration features. It complements response alignment: response alignment guides where each expert responds, while HSIC regularizes what each expert represents.

\section{Experiments}

In this section, we evaluate CoRe-UIE on representative underwater image enhancement benchmarks. We conduct full-reference comparisons on UIEB and LSUI, no-reference generalization evaluation on U45, qualitative comparisons, and ablation studies to verify the effectiveness of the proposed components.

\subsection{Experimental Setup}

\begin{figure*}[t]
    \centering
    \includegraphics[width=1\linewidth]{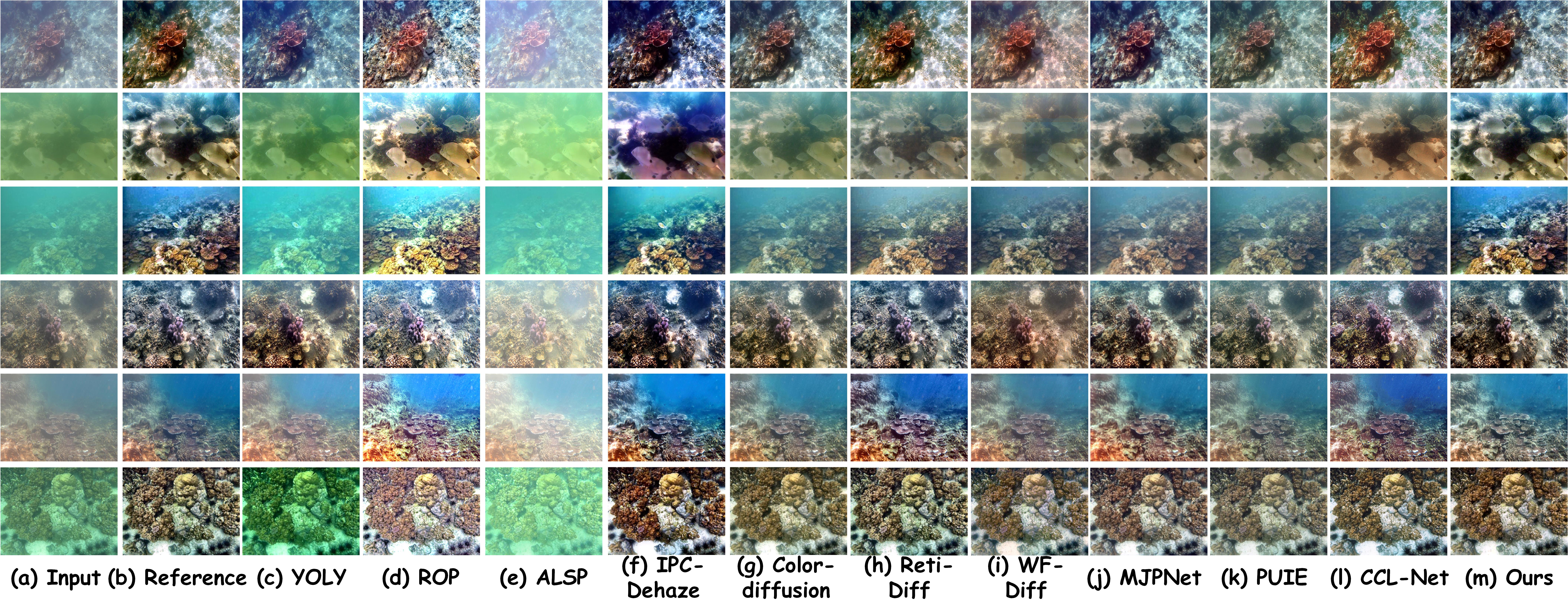}
    \caption{Qualitative comparison of underwater image enhancement results.}
    \label{fig:qualitative}
\end{figure*}

\textbf{Datasets and Baselines.}
We evaluate CoRe-UIE on UIEB~\cite{li2019uieb}, LSUI~\cite{peng2023ushape}, and U45~\cite{li2019fgan}. UIEB and LSUI provide reference images and are used for full-reference evaluation. U45 contains real underwater images without ground-truth references, and is used to evaluate the generalization ability of different methods under no-reference metrics. We compare CoRe-UIE with representative underwater enhancement methods, including IPC-Dehaze~\cite{fu2025ipcdehaze}, Color-Diffusion~\cite{chang2026diffcolor}, Reti-Diff~\cite{he2025retidiff}, WF-Diff~\cite{zhao2024wfdiff}, MJPNet~\cite{fan2025mjpnet}, PUIE~\cite{fu2022puie}, CCL-Net~\cite{liu2024cclnet}, YOLY~\cite{li2021yoly}, ROP~\cite{liu2021rop},  ALSP~\cite{he2025alsp}, SS-UIE~\cite{ peng2025adaptive}, PMGMamba~\cite{tan2026pgmamba}.

\textbf{Implementation Details.}
CoRe-UIE is implemented in PyTorch and trained on a Nvidia RTX PRO 6000 GPU. During training, input images are randomly cropped into fixed-size patches and augmented with random horizontal flipping. Adam is adopted as the optimizer with \(\beta_1=0.9\) and \(\beta_2=0.999\). The initial learning rate is set to \(3\times10^{-6}\), and gradient accumulation is used when necessary. The shared expert is activated for all regions, while the region-adaptive Top-\(k\) router selects the most relevant mechanism experts for local restoration.

\textbf{Evaluation Metrics.}
For UIEB and LSUI, we adopt PSNR, SSIM~\cite{wang2004ssim}, FSIM/FSIMC~\cite{zhang2011fsim}, and VSI~\cite{zhang2014vsi} for full-reference evaluation. Since U45 does not provide ground-truth references, we further employ no-reference metrics, including NIQE~\cite{mittal2013niqe}, BRISQUE~\cite{mittal2012brisque}, CEIQ~\cite{yan2019ceiq}, and PIQE~\cite{venkatanath2015piqe}. For PSNR, SSIM, FSIM, FSIMC, VSI, and CEIQ, higher values indicate better quality, while lower NIQE, BRISQUE, and PIQE scores indicate better perceptual quality.

\subsection{Experimental Results}

\textbf{Full-Reference Evaluation.}
Table~\ref{tab:comparison} reports the quantitative results on UIEB and LSUI. On UIEB, CoRe-UIE achieves the best PSNR, SSIM, FSIM, FSIMC, and VSI scores, reaching 26.88 dB, 0.9103, 0.9666, 0.9554, and 0.9855, respectively. Compared with diffusion-based methods such as Reti-Diff and WF-Diff, CoRe-UIE obtains higher fidelity and structural consistency. On LSUI, CoRe-UIE also achieves the best results across all metrics, outperforming strong competitors such as WF-Diff and PUIE. These results indicate that decomposing underwater restoration into color correction, scattering suppression, texture recovery, and illumination protection is effective under different degradation distributions.

\begin{table*}[t]
\centering
\small
\setlength{\tabcolsep}{4pt}
\begin{tabular}{clccccc}
\toprule
Dataset & Method & PSNR(dB)$\uparrow$ & SSIM$\uparrow$ & FSIM$\uparrow$ & FSIMC$\uparrow$ & VSI$\uparrow$ \\
\midrule
\multirow{11}{*}{UIEB}
& YOLY & 13.7816 & 0.4509 & 0.7966 & 0.7682 & 0.9170 \\
& ROP & 20.2698 & 0.7467 & 0.8626 & 0.8406 & 0.9401 \\
& ALSP & 12.7916 & 0.7416 & 0.9011 & 0.8674 & 0.9458 \\
& IPC-Dehaze & 19.4835 & 0.8315 & 0.9450 & 0.9297 & 0.9720 \\
& Color-Diffusion & 24.7674 & 0.8044 & 0.9617 & 0.9480 & 0.9791 \\
& Reti-Diff & 26.4449 & 0.8377 & 0.9548 & 0.9530 & 0.9821 \\
& WF-Diff & 25.9754 & 0.8603 & 0.9457 & 0.9268 & 0.9704 \\
& MJPNet & 25.0001 & 0.7562 & 0.9339 & 0.9206 & 0.9702 \\
& PUIE & 24.7498 & 0.8994 & 0.9611 & 0.9489 & 0.9798 \\
& CCL-Net & 22.2415 & 0.8944 & 0.9603 & 0.9460 & 0.9797 \\
& PGMamba & 24.5684 & 0.8992 & 0.9562 & 0.9493 & 0.9738 \\
& SSUIE & 22.5749  & 0.8713  & 0.9462 & 0.9339 & 0.9613\\
& Ours & \textbf{26.8815} & \textbf{0.9103} & \textbf{0.9666} & \textbf{0.9554} & \textbf{0.9855} \\
\midrule
\multirow{11}{*}{LSUI}
& YOLY & 13.6900 & 0.6203 & 0.8254 & 0.7212 & 0.9210 \\
& ROP & 13.8153 & 0.6520 & 0.8162 & 0.7934 & 0.9206 \\
& ALSP & 12.0710 & 0.6889 & 0.8906 & 0.8354 & 0.9240 \\
& IPC-Dehaze & 19.4625 & 0.7660 & 0.9083 & 0.8869 & 0.9563 \\
& Color-Diffusion & 20.8894 & 0.5842 & 0.8866 & 0.8711 & 0.9538 \\
& Reti-Diff & 22.1464 & 0.8026 & 0.9134 & 0.8998 & 0.9652 \\
& WF-Diff & 22.1320 & 0.8359 & 0.9303 & 0.9103 & 0.9665 \\
& MJPNet & 20.7294 & 0.7920 & 0.9093 & 0.8956 & 0.9642 \\
& PUIE & 21.8375 & 0.8291 & 0.9288 & 0.9212 & 0.9699 \\
& CCL-Net & 20.3836 & 0.8076 & 0.9230 & 0.9065 & 0.9647 \\
& PGMamba & 21.96 & 0.8246 & 0.9279 & 0.9043 & 0.9537 \\
& SSUIE & 21.5421  & 0.8167  & 0.9023 & 0.8926 & 0.9058\\
& Ours & \textbf{22.2997} & \textbf{0.8452} & \textbf{0.9366} & \textbf{0.9230} & \textbf{0.9716} \\
\bottomrule
\end{tabular}
\caption{Quantitative comparison on UIEB and LSUI datasets.}
\label{tab:comparison}
\end{table*}\

\textbf{No-Reference Generalization on U45.}
Since U45 does not provide ground-truth references, we evaluate no-reference generalization using NIQE, BRISQUE, CEIQ, and PIQE. As shown in Table~\ref{tab:u45_nr}, CoRe-UIE achieves the best results on all four metrics, suggesting favorable perceptual quality and robustness on reference-free underwater scenes.

\begin{table}[t]
\centering
\small
\setlength{\tabcolsep}{3pt}
\begin{tabular}{lcccc}
\toprule
Method & NIQE$\downarrow$ & BRISQUE$\downarrow$ & CEIQ$\uparrow$ & PIQE$\downarrow$ \\
\midrule
YOLY & 4.3429 & 21.2950 & 3.0938 & 23.2958 \\
ROP & 4.6938 & 21.2902 & 3.4422 & 25.4535 \\
ALSP & 3.5711 & 21.0875 & 2.6285 & 17.4100 \\
IPC-Dehaze & 4.0849 & 20.7812 & 3.4714 & 25.1462 \\
Color-Diffusion & 4.2948 & 21.3285 & 3.1938 & 19.9857 \\
Reti-Diff & 3.8254 & 18.1158 & 3.4786 & 18.9045 \\
WF-Diff & 3.9074 & 17.4601 & 3.3394 & 15.7289 \\
MJPNet & 4.2917 & 23.6033 & 3.4671 & 15.6447 \\
PUIE & 3.8673 & 20.1637 & 3.2959 & 19.6707 \\
CCL-Net & 3.7991 & 19.1456 & 3.3834 & 19.0426 \\
PGMamba & 3.5737 & 18.6443 & 3.3671 & 15.7649 \\
SSUIE & 3.6428  & 19.2167  & 3.2439 & 16.5716 \\
Ours & \textbf{3.4874} & \textbf{17.4573} & \textbf{3.4846} & \textbf{15.4892} \\
\bottomrule
\end{tabular}
\caption{No-reference quantitative comparison on U45.}
\label{tab:u45_nr}
\end{table}

\begin{figure}[t]
    \centering
    \includegraphics[width=1\linewidth]{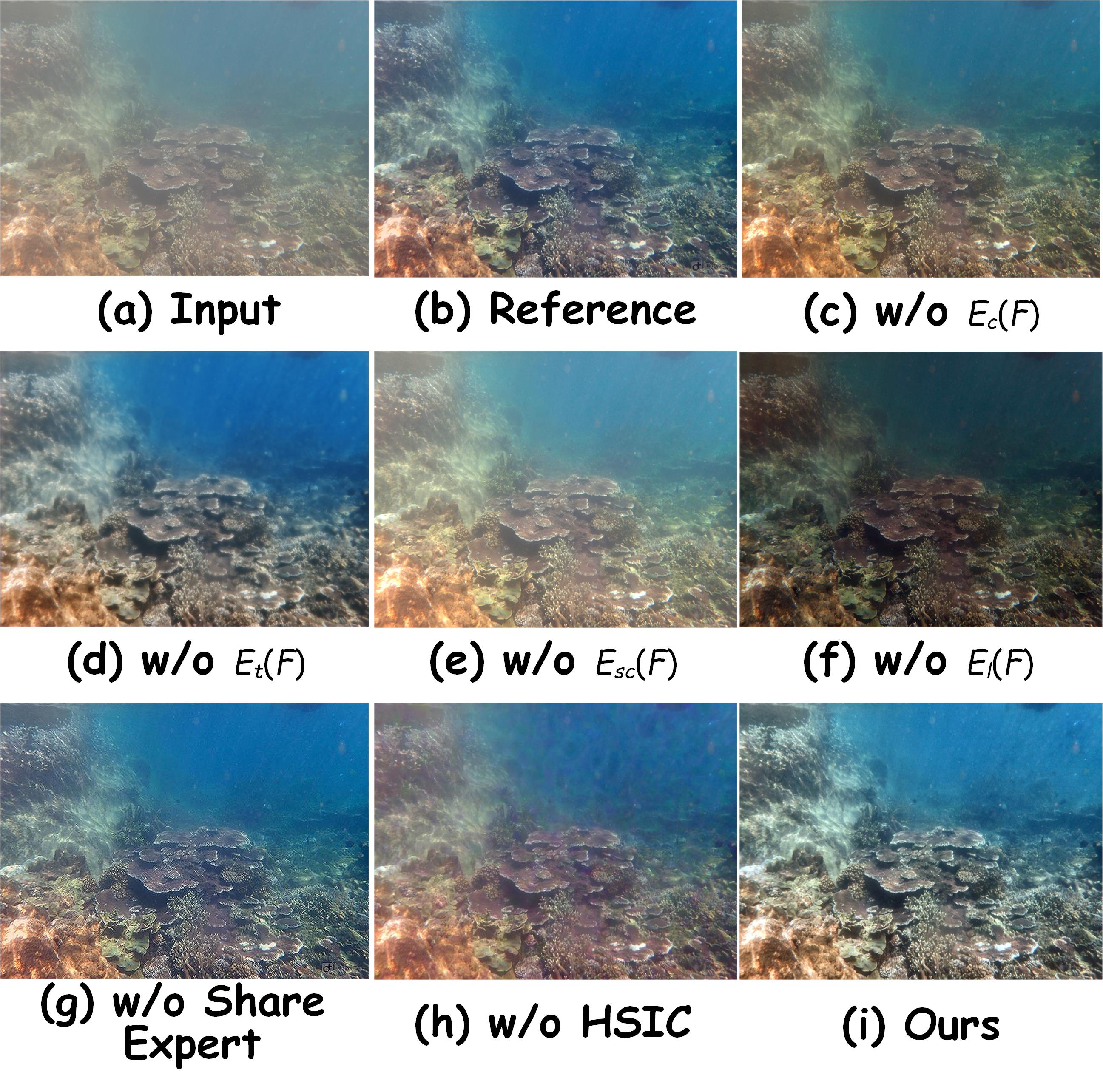}
    \caption{Qualitative ablation study of mechanism-guided experts and the HSIC constraint.}
    \label{fig:ablation_expert}
\end{figure}

\begin{figure}[t]
    \centering
    \includegraphics[width=1\linewidth]{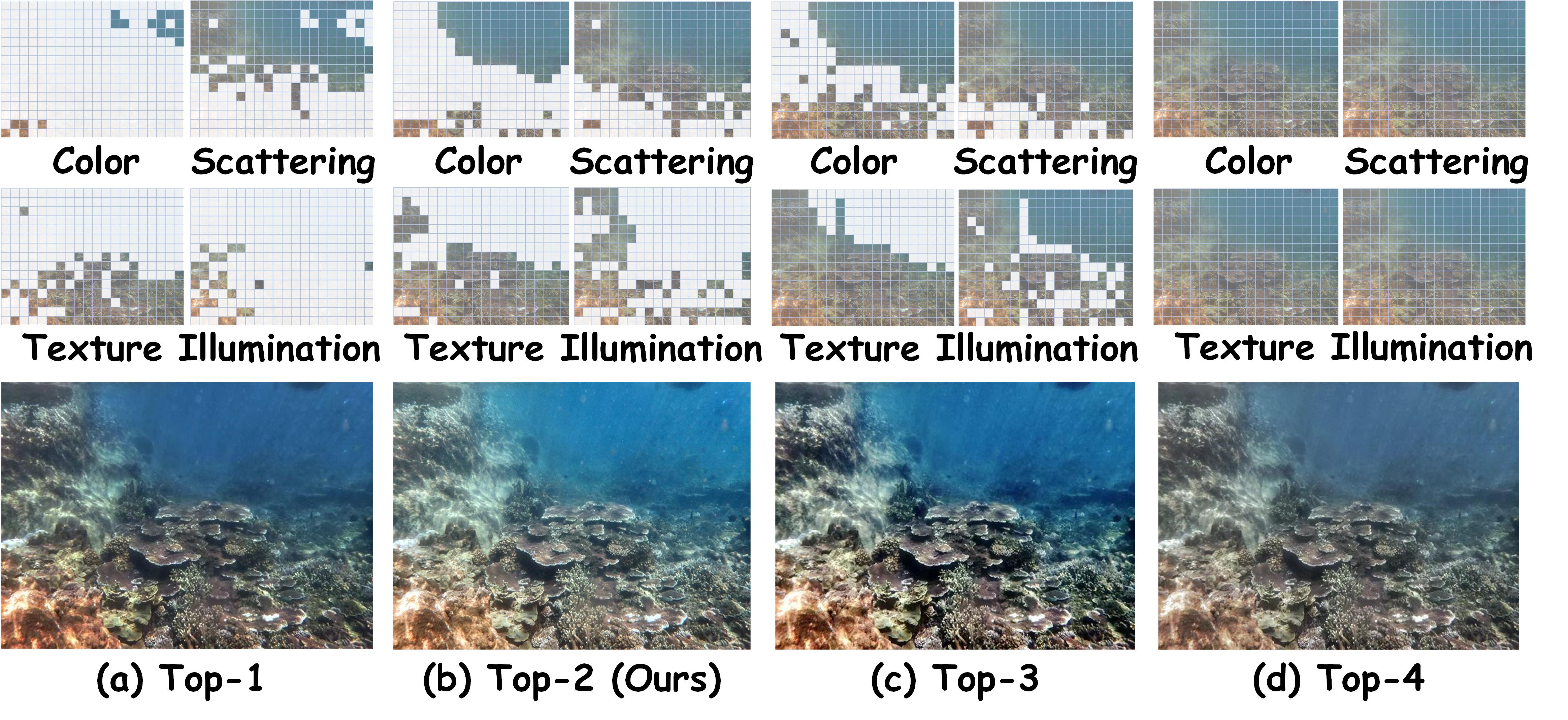}
    \caption{Qualitative comparison of different Top-\(k\) routing settings.}
    \label{fig:ablation_topk}
\end{figure}

\textbf{Qualitative Analysis.}
Figure~\ref{fig:qualitative} presents visual comparisons among different methods. IPC-Dehaze, YOLY, and ROP can improve visibility in some cases, but they often leave residual color casts or produce unstable contrast. Color-Diffusion and Reti-Diff generate stronger color enhancement, yet their results may exhibit over-saturation or local color shifts. WF-Diff improves global visibility through frequency-domain diffusion, but some fine textures and local colors remain less consistent with the reference. PUIE produces relatively natural results in several scenes, but may still retain haze or weak local details. CCL-Net, MJPNet, and ALSP provide balanced enhancement in some cases, but may suffer from insufficient haze removal or illumination artifacts. In contrast, CoRe-UIE produces more consistent visual results with natural color appearance, clearer structures, and fewer over-enhancement artifacts. This is because different local regions are restored by different combinations of mechanism-specific experts instead of being processed through a single shared restoration pathway.

\subsection{Ablation Study}

\begin{table}[t]
\centering
\small
\setlength{\tabcolsep}{3.5pt}
\begin{tabular}{cccccc}
\toprule
 Method & PSNR & SSIM & FSIM & FSIMC & VSI \\
\hline
w/o \(E_c(F)\) & 26.0342 & 0.8933 & 0.9226 & 0.9198 & 0.9650 \\
w/o \(E_t(F)\) & 26.5769 & 0.8819 & 0.9448 & 0.9229 & 0.9710 \\
w/o \(E_{sc}(F)\) & 25.8933 & 0.9075 & 0.8916 & 0.8672 & 0.9623 \\
w/o \(E_l(F)\) & 26.3260 & 0.8876 & 0.9223 & 0.9213 & 0.9755 \\
w/o Shared Expert & 26.4795 & 0.8931 & 0.9587 & 0.9331 & 0.9798 \\
w/o HSIC & 26.0224 & 0.9018 & 0.9210 & 0.9207 & 0.9694 \\
\midrule
Top-1 & 26.7944 & 0.9018 & 0.9429 & 0.9307 & 0.9670 \\
Top-3 & 26.3138 & 0.9007 & 0.9523 & 0.9295 & 0.9556 \\
Top-4 & 26.6910 & 0.8991 & 0.9305 & 0.9374 & 0.9427 \\
Ours(ToP-2) & \textbf{26.8815} & \textbf{0.9103} & \textbf{0.9666} & \textbf{0.9554} & \textbf{0.9821} \\
\bottomrule
\end{tabular}
\caption{Ablation study of mechanism-guided experts and different Top-\(k\) routing choices. The best results are highlighted in bold.}
\label{tab:ablation}
\end{table}

\textbf{Effect of Mechanism-Guided Experts.}
As shown in Table~\ref{tab:ablation}, removing any routed expert leads to performance degradation, indicating that shared-backbone experts can learn complementary restoration roles under the proposed routing and regularization strategy. Removing the color-related expert weakens color compensation, while removing the scattering-related expert causes the largest drop, confirming the importance of visibility-related correction. Removing the texture- or illumination-related expert also reduces structural quality and brightness stability. The variant without the shared expert performs worse, suggesting that mechanism-invariant content information is useful for stable reconstruction.

The variant without HSIC also degrades performance. Since all routed experts adopt the same backbone architecture, removing HSIC makes expert representations more likely to overlap under the shared reconstruction objective. This verifies the importance of HSIC-based representation disentanglement for maintaining expert complementarity.

\textbf{Effect of Top-\(k\) Routing.}
We further compare Top-1, Top-2, Top-3, and Top-4 routing. Top-1 provides strong selectivity but may miss complementary restoration mechanisms. Top-3 and Top-4 activate more experts, but their denser routing weakens expert selectivity and may introduce redundant corrections. Top-2 achieves the best balance between collaboration and selectivity, and is therefore adopted as the default setting.Therefore, we adopt Top-2 as the default setting in CoRe-UIE, as it provides a favorable trade-off between restoration accuracy and computational efficiency by preserving necessary expert collaboration while avoiding redundant dense expert activation.
\section{Conclusion}

We present mechanism-aware adaptive restoration framework CoRe-UIE for underwater image enhancement. Instead of relying on a unified restoration pathway, CoRe-UIE uses shared-backbone routed experts with region-adaptive Top-\(k\) routing to handle spatially diverse underwater degradations. A content-preserving shared expert provides mechanism-invariant restoration information, while routed experts learn degradation-related correction behaviors under mechanism-cue-guided response alignment and HSIC-based representation disentanglement. Experimental results on multiple underwater benchmarks demonstrate that CoRe-UIE achieves competitive quantitative performance and visually balanced enhancement, especially for scenes with coexisting color distortion, scattering haze, texture degradation, and illumination variation.

\bibliography{aaai2027}


\end{document}